\documentclass{article}

\usepackage{arxiv}

\usepackage[utf8]{inputenc} 
\usepackage[T1]{fontenc}    
\usepackage{hyperref}       
\usepackage{url}            
\usepackage{booktabs}       
\usepackage{amsfonts}       
\usepackage{nicefrac}       
\usepackage{microtype}      
\usepackage{lipsum}		
\usepackage{graphicx}
\usepackage[numbers]{natbib}
\usepackage{doi}
\usepackage{crop}%
\usepackage{amsmath,amssymb,amsfonts,upref,endnotes}
\usepackage{amsthm}
\usepackage{marginnote}
\usepackage{soul}
\usepackage{amsmath}
\usepackage[figuresright]{rotating}
\usepackage{color}

\newcommand*{\permcomb}[4][0mu]{{{}^{#3}\mkern#1#2_{#4}}}

\newcommand*{\comb}[1][-1mu]{\permcomb[#1]{C}}

\title{Adaptive Sensor Steering Strategy Using Deep Reinforcement Learning for Dynamic Data Acquisition in Digital Twins}

\author{ \href{https://orcid.org/0000-0002-3672-0240}{\includegraphics[scale=0.06]{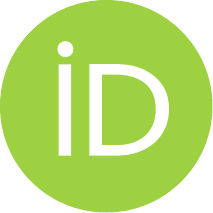}\hspace{1mm}Collins O.~Ogbodo} \\
	Dynamic Research Group\\
	School of Mechanical, Aerospace and Civil Engineering\\
        University of Sheffield\\
	Mapping Street, Sheffield, S1 3JD, United Kingdom. \\
	\texttt{coogbodo1@sheffield.ac.uk} \\
	\And
        \href{https://orcid.org/0000-0002-3433-3247}{\includegraphics[scale=0.06]{orcid.pdf}\hspace{1mm}Timothy J. ~Rogers} \\
	Dynamic Research Group\\
	School of Mechanical, Aerospace and Civil Engineering\\
        University of Sheffield\\
	Mapping Street, Sheffield, S1 3JD, United Kingdom. \\
	\texttt{Tim.rogers@sheffield.ac.uk} \\
        \And
        \href{https://orcid.org/0000-0003-4263-0513}{\includegraphics[scale=0.06]{orcid.pdf}\hspace{1mm}Mattia Dal ~Borgo} \\
	Siemens Digital Industries Software NV\\
	Interleuvenlaan 68, 3001 Leuven, Belgium.\\
	\texttt{mattia.dal\_borgo@siemens.com} \\
        \And
	\href{https://orcid.org/0000-0002-7266-2105}{\includegraphics[scale=0.06]        {orcid.pdf}\hspace{1mm}David J. ~Wagg} \\
	Dynamic Research Group\\
	School of Mechanical, Aerospace and Civil Engineering\\
        University of Sheffield\\
	Mapping Street, Sheffield, S1 3JD, United Kingdom. \\
        The Alan Turing Institute\\
        NW1 2DB, London, United Kingdom.\\
	\texttt{David.wagg@sheffield.ac.uk} \\
}

\renewcommand{\shorttitle}{Adaptive Sensor Steering Strategy.}

\hypersetup{
pdftitle={Adaptive Sensor Steering Strategy Using Deep Reinforcement Learning for Dynamic Data Acquisition in Digital Twins},
pdfsubject={q-bio.NC, q-bio.QM},
pdfauthor={David S.~Hippocampus, Elias D.~Striatum},
pdfkeywords={Digital Twins, Reinforcement Learning, Information Theory, Design of Experiment, Structural Health Monitoring.},
}

\begin{document}
\maketitle

\begin{abstract}
This paper introduces a sensor steering methodology based on deep reinforcement learning to enhance the predictive accuracy and decision support capabilities of digital twins by optimising the data acquisition process. Traditional sensor placement techniques are often constrained by one-off optimisation strategies, which limit their applicability for online applications requiring continuous informative data assimilation. The proposed approach addresses this limitation by offering an adaptive framework for sensor placement within the digital twin paradigm. The sensor placement problem is formulated as a Markov decision process, enabling the training and deployment of an agent capable of dynamically repositioning sensors in response to the evolving conditions of the physical structure as represented by the digital twin. This ensures that the digital twin maintains a highly representative and reliable connection to its physical counterpart. The proposed framework is validated through a series of comprehensive case studies involving a cantilever plate structure subjected to diverse conditions, including healthy and damaged conditions. The results demonstrate the capability of the deep reinforcement learning agent to adaptively reposition sensors improving the quality of data acquisition and hence enhancing the overall accuracy of digital twins.
\end{abstract}


\section{Introduction}
The increasing complexity and multidisciplinary nature of engineering systems over the years has continuously motivated the development of new methods for analysis, monitoring, and optimisation. This increase has led to the emergence of digital technologies such as digital twins --- a virtual replica of a physical asset that enables a bidirectional coupling between digital and physical spaces \cite{wagg2025philosophical}. The digital twin paradigm provides opportunities across different engineering disciplines, including technical risk reduction and reduced cost of experiments in digital engineering, enhanced first-time yield and product optimisation in manufacturing, and improved system capability and operational availability in operation engineering \cite{ferrari2024digital}. In other fields like precision medicine, digital twins are utilised for patient-personalised medical care \cite{corral2020digital} and in geophysical science for estimation of ice sheet flow model parameters \cite{isaac2015scalable}. Digital twin systems are built from the integration of models and data using sophisticated algorithms, expert know-how, and digital connectivity, presenting significant predictive capability when compared to current technologies \cite{wagg2020digital}. Within engineering systems, this integration involves the combination of physics-based and data-driven models for improved reliability of predictive results. Critical aspects such as continual validation and updating ensure that the virtual twin maintains a good representation of the physical system. To achieve this representation, digital twin systems assimilate streams of data from the physical asset in real or near real-time. 

While the decreasing cost of sensors and enhanced connectivity has made available a large amount of data, the quality of data assimilated by the digital twin is a key determinant of its predictive accuracy. Given this dependence, data used in the digital twin update process are expected to contain relevant patterns of interest, be characterised by low noise levels, and reproducible under the same circumstances \cite{cicirello2024physics}. To ensure the virtual representation is tailored to the behaviour of its physical counterpart, a critical part of the virtual-to-physical feedback flow of a digital twin is a dynamic data acquisition process. In scenarios where data are scarce, methods such as active learning and reinforcement learning can guide the collection of additional data that is most relevant to the digital twin's objectives \cite{willcox2024role}.

This work therefore presents a novel approach for the dynamic data acquisition process of digital twins using a sensor steering methodology based on deep reinforcement learning (DRL). Our proposed framework learns a sensor steering policy that is deployed online within the virtual space and decides on data acquisition strategies, particularly sensor location, to maximise the information content of acquired data. This approach extends the adaptability of digital twins to the changing behaviours of the physical asset \cite{willcox2024role}. 

This paper is organised as follows: Section \hyperref[sec:Background]{2} presents a review of previous related methodologies of sensor placement and their respective limitations within the digital twin paradigm. In Section \hyperref[sec:Problem Formulation]{3}, the problem definition and mathematical abstractions via DRL are provided. Section \hyperref[sec:Proposed Framework]{4}  establishes how the proposed methodology integrates into the digital twin paradigm. Section \hyperref[sec:Results]{5} presents the results of the proposed strategy when applied to a cantilever plate structure. Finally, Section \hyperref[sec:Conclusion]{6} summarises our main contributions, including limitations and future directions.

\section{Background}\label{sec:Background}
The problem of optimal sensor placement (OSP) has been well researched across multiple studies within structural dynamics \cite{hassani2023systematic, tan2020computational}. The key aspects of OSP research are the evaluation criteria and optimisation algorithms \cite{wang2023advancements}. The evaluation criteria align with specific monitoring objectives and effectively reflect the quality of the sensor placement solution, and the iterative search for optimal solutions is largely influenced by the selected criteria's structure, which significantly impacts computational efficiency\cite{wang2023advancements}. Different evaluation criteria have been extensively explored, some of which aim to optimise a prespecified objective, such as maximising a characteristic vibration signal, typically kinetic energy, which has been found to yield sub-optimal results due to its independence on the parameters being estimated \cite{udwadia1994methodology}, modal kinetic energy \cite{salama1987optimal, meo2005optimal} and driving point residues \cite{chang2014optimal, li2012optimal}. Other criteria are based on maximising modal identification metrics, like modal assurance criteria \cite{yang2021adaptive} and singular value decomposition ratio \cite{wang2014sensor}, as well as information-based criteria such as information entropy \cite{papadimitriou2000entropy} and mutual information \cite{bhattacharyya2020exploiting}. A summary of these criteria is presented in Wang et al\cite{wang2023advancements} with their respective merits and demerits. 
 The optimisation algorithm, on the other hand, is the computational strategy adopted to either minimise or maximise the selected evaluation criteria, which is usually iterative. A number of optimisation algorithms have also been explored. Some of these are evolutionary algorithms, notably genetic algorithms, particle swarm optimisation, and simulated annealing \cite{tan2020computational}. Other heuristic algorithms such as the forward and backward sequential sensor placement (FSSP and BSSP) have also been successfully applied to obtain optimal locations \cite{papadimitriou2012effect}. Recent research has also explored data-driven approaches based on DRL within an optimisation formulation to search the placement space for optimal location. DRL was applied in Jabini \& Johnson\cite{jabini2022deep} to find the optimal location of two sensors in a base-isolated system with five degrees of freedom using a value-based deep Q-Network agent.
 
\begin{figure}[htbp]
  \centering
    \includegraphics[width=1\textwidth]{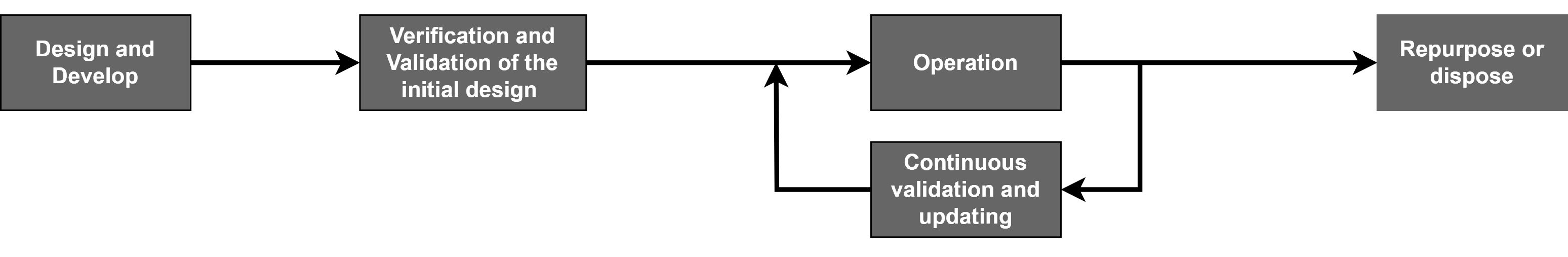} 
  \caption{The digital twin life cycle which involves continual updating and validation to maintain accuracy in prediction and decision support.}
  \label{fig:Digital Twin Life cycle}
\end{figure}

 Within the digital twin paradigm, where data assimilation and continuous updating are required, acquiring high-quality data is essential to ensure the digital twin maintains an accurate representation of the physical asset \cite{wagg2020digital}. An essential stage of the digital twin life cycle is continuous validation against operational data, which ensures accuracy in prediction and decision support. This process is highly dependent on the data acquired at every stage of the operational lifespan of the physical asset (Figure \ref{fig:Digital Twin Life cycle}). While existing sensor placement strategies are suitable for traditional structural monitoring operations and focus on finding a single optimal solution that remains fixed throughout the entire lifespan of the structure, the digital twin framework typically requires an adaptive approach. This approach must consider the changing dynamic behaviour of the physical structure and align sensor locations to guarantee the continuous acquisition of high-quality data. This adaptive sensing strategy was researched in Kapteyn \& Willcox \cite{kapteyn2022design}, which presented an interpretable approach based on an optimal decision tree. The authors proposed a sensor scheduling methodology that was applied on a mathematical abstraction of an unmanned aerial vehicle (UAV) digital twin based on a model library of damages \cite{kapteyn2021probabilistic}. However, this strategy assumes that the damage states of the structure are known ab initio and therefore has a distribution of sensors across the wing of the UAV, which results in the deployment of a significant number of sensors.

\begin{figure}[htbp]
  \centering
    \includegraphics[width=1\textwidth]{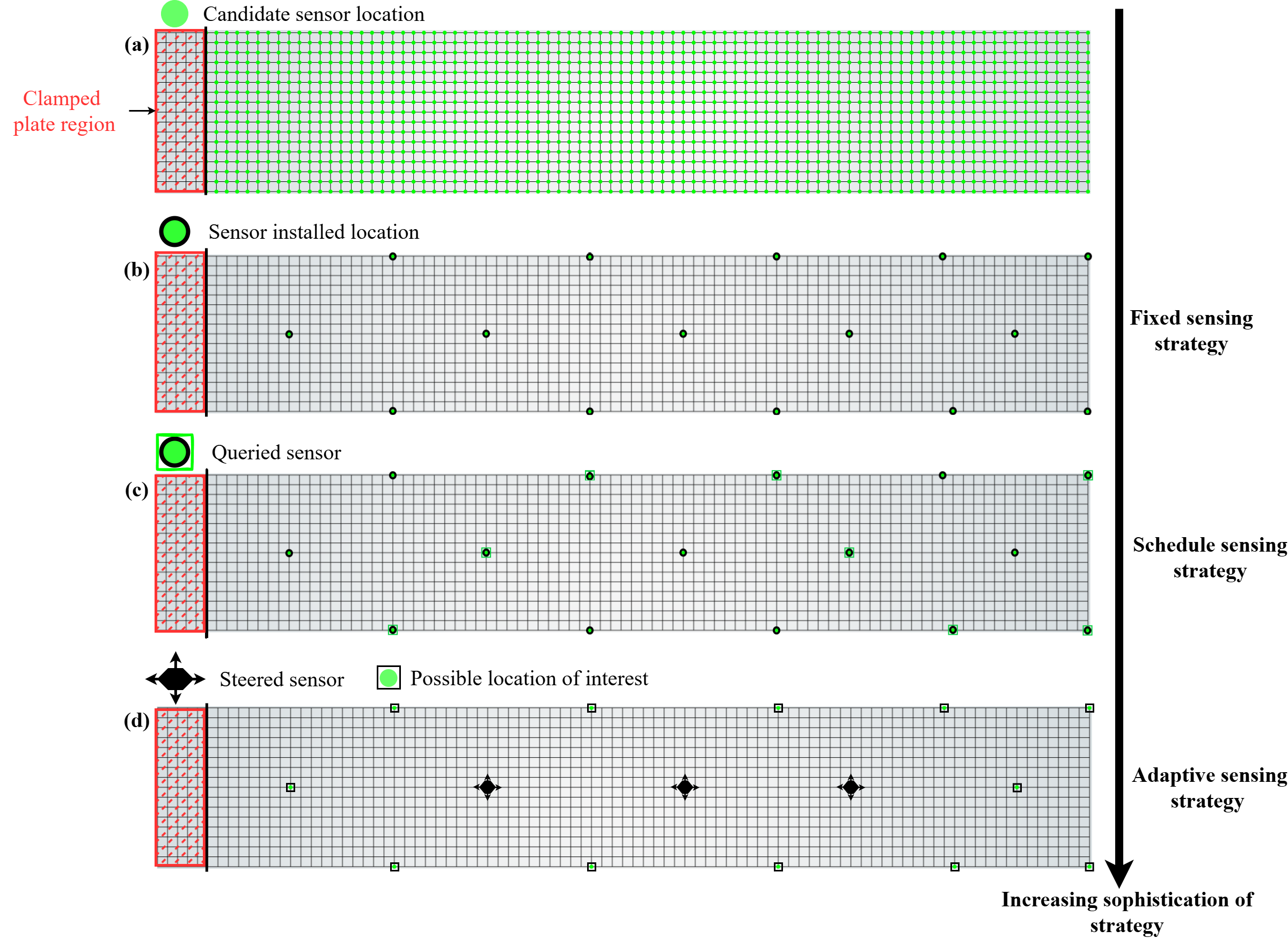} 
    \caption{Evolution of sensor placement strategies for the example of a rectangular plate showing (a) a grid of all possible sensor candidate locations, (b) fixed sensing strategy-- sensors locations are predefined and remain fixed throughout the structure's lifespan, (c) sensor scheduling strategy-- sensors locations are predefined and queried at different times, and (d) sensor steering strategy-- sensors are adapted to new locations of interest across the lifecycle of the structure.}
  \label{Fig: Evolution of sensing strategy}
\end{figure}

In this study, we shift from a sensor scheduling approach to a sensor steering approach as shown in Figure \ref{Fig: Evolution of sensing strategy} for the example of a rectangular plate. Sensor steering involves a virtual-to-physical change \cite{ferrari2024digital}, which offers significant benefits for monitoring structures and further enhancing control. Unlike sensor scheduling, which relies on premounted sensors, sensor steering dynamically adjusts sensor positions based on real-time or near real-time data and evolving conditions of the physical asset. This approach ensures that the digital twin continuously receives the most relevant data. The enhanced methodology provides a robust framework for adaptive sensor placement within the digital twin paradigm, ensuring that the digital twin remains an accurate and reliable representation of the physical asset throughout its operational lifespan.

Our contribution therefore differs from previous work in three aspects:
\begin{itemize}
    \item While previous work presented above is either fixed or scheduling strategies, ours is an adaptive steering strategy that does not involve computing the model's time response for training data.
    \item Although reinforcement learning has been explored previously within an optimisation setting, we follow a policy-based approach that is suitable for online decision processes such as digital twins and other data-driven models.
    \item Our strategy employs a distributional reinforcement learning framework, presenting opportunities for placement decisions based not only on the expected return but also on alternative metrics such as variance and conditional value at risk (CVaR) of the return distribution.
\end{itemize}

\section{Problem formulation}\label{sec:Problem Formulation}
The primary objective of this study is to develop a sensor steering strategy that dynamically adapts the location of sensors on a structure in response to changing conditions. For this, we adopt an agent-based approach where we learn how to act (change the sensor position) given the current state of the physical asset as represented by the digital twin. This section presents a mathematical formulation of the problem using the Markov decision process (MDP) and its implementation with DRL.

\subsection{Markov decision process formulation}
\begin{figure}[htbp]
  \centering
    \includegraphics[width=1\textwidth]{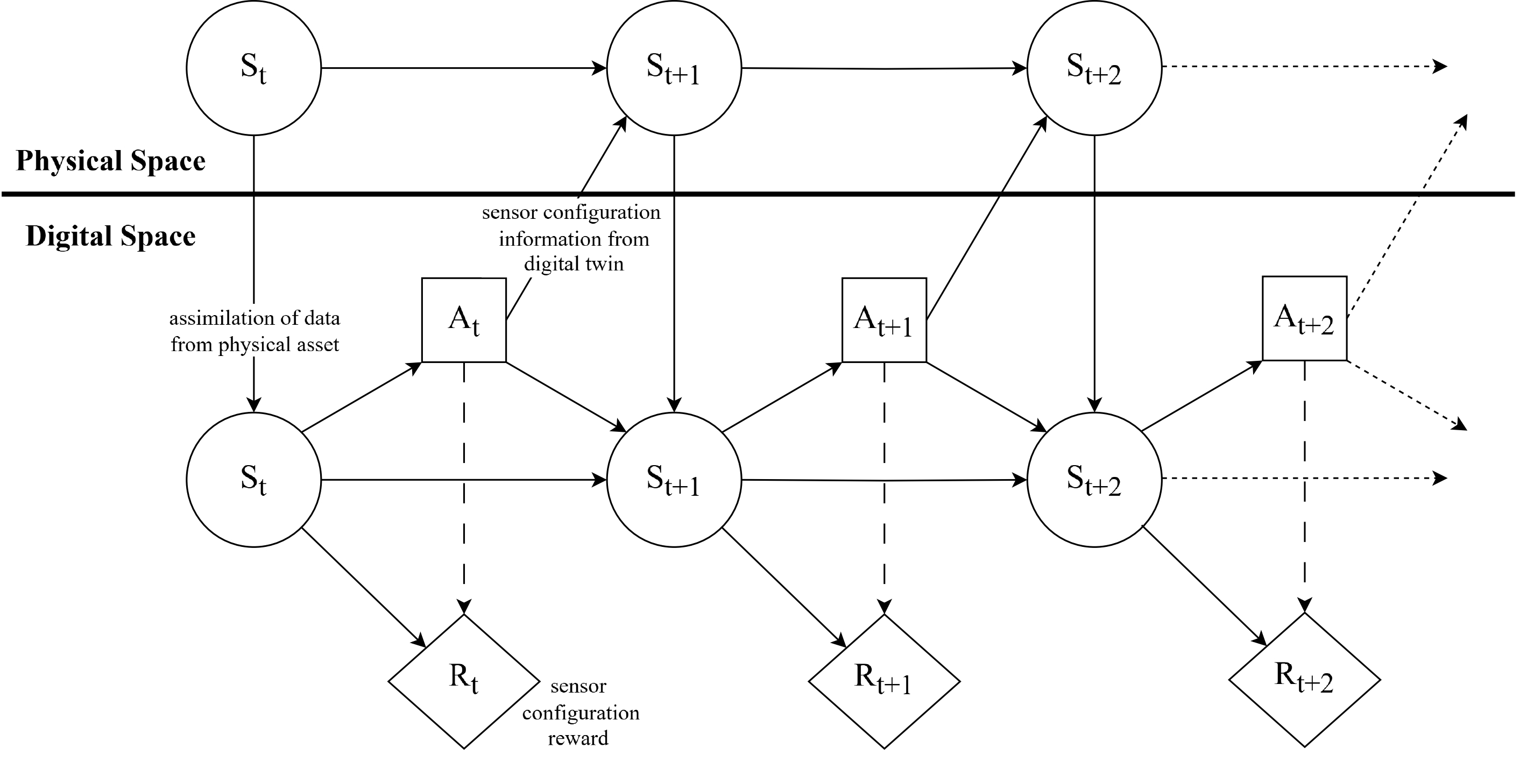} 
  \caption{Graphical model of dynamic decision interaction between physical and digital space showing sensor steering information exchange.}
  \label{Fig: Grpahical model}
\end{figure}

The sensor steering problem can be formulated as a sequential decision-making process, where the goal is to determine the optimal sensor configuration at each time step. This involves deciding which sensors to move and in which direction, based on the current state of the digital twin. At each step, the sensor configuration is dependent only on the selected sensor and the steering direction, which can be modelled with the MDP formulation defined by the tuple $\langle \mathcal{S}, \mathcal{A}, \mathcal{P}, \mathcal{R}, \gamma \rangle$ where 
\begin{itemize}
    \item $s \in \mathcal{S}$ is the set of states representing the possible sensor configurations.
    \item $a \in \mathcal{A}$ is the set of actions containing all possible sensor motion.
    \item $\mathcal{P}: p(s' | s,a) := Pr\{S_{t+1} =s' | S_{t} = s, A_{t} = a\} $ is the state transition probability matrix, representing the dynamics of the virtual environment.
    \item $\mathcal{R}: R(s, a)$ is the reward function, which quantifies the quality of the sensor configuration.
    \item $\gamma \in [0,1]$ is the discount factor, which determines the importance of the final sensor location rewards.
\end{itemize}
The problem also naturally satisfies the Markov assumption, denoted as
\begin{equation}
    Pr\{S_{t+1} = s' | S_t = s, A_t =a \} = Pr\{S_{t+1} = s' | S_t = s_{t, t-1,...,0}, A_t =a \}
    \label{equ:Markov Assumption}
\end{equation}
where $s_{t, t-1,...,0}$ is the state trajectory history. A graphical model of the process similar to that developed by Tezzele et al.\cite{tezzele2024adaptive} is shown in Figure  \ref{Fig: Grpahical model}. The graphical model shows the physical-to-virtual and virtual-to-physical data and information interaction.

\subsubsection{Action space}
A multidiscrete action space \cite{towers2024gymnasium} is used in the form of a vector of discrete numbers with length equal to two actions: (1) selected sensor $p \in \mathbf{P}$ bounded by the maximum number of sensors specified in the environment setup; (2) steering direction $g \in \mathbf{G}$ bounded by the motion directions ($g_0$ =left, $g_1$= right, $g_2$= up and $g_3$ = down). Actions are sampled from a space of $\mathbb{R}^{\mathbf{P} \times \mathbf{G}}$ and executed in the environment. A third action which encoded the number of steps to take in a specific direction was originally considered, but this significantly increased the dimension of the action space, which reduced the efficiency of learning and hence was not used in this study. The action space is further reduced to a discrete action to enable the use of a value-based learning algorithm. Therefore, a typical action $a = 1$ is interpreted within the environment as $[0,1]$ which means move sensor $p_0$ one step in the right direction $g_1$.

\subsubsection{State space}
The State is defined by a multi-binary space \cite{towers2024gymnasium} of dimension $\mathbb{B}^{n^*}$ where $\mathbb{B} = \{0,1\}$ and $n^*$ is the number of mesh nodes in the pre-specified placement region of the discretised model. The space is represented by a vector with length $n^*$. Nodes represent possible sensor candidate positions and are encoded with ones when instrumented with a sensor and zeroes otherwise. Each sensor position combination represents a state of the environment. 
 
\subsection{Reward function}\label{subsec:Reward Function}
Considering that the sensor steering problem is fundamentally built on an OSP formulation, which is in itself a combinatoric problem that involves the search of $u$ candidate position that maximises a specified objective function given $v$ possible position, where $u < v$. The space of possible sensor combinations is then given as $\comb{v}{u}$ which scales up significantly with the size and complexity of the structure. Unlike OSP, which involves running the underlying optimisation once to produce an optimal combination result, the sensor steering process is a learning problem that results in a policy and therefore is more computationally intensive, ideally requiring a fast, easy-to-compute objective function. Given the nature of the learning problem, an information-theoretic objective function satisfies our learning requirement, particularly the Fisher information matrix (FIM) metric.

\subsubsection{Information theoretic reward}
The learning goal, by extension, is to acquire highly informative data, which makes an information entropy reward function suitable. Information entropy serves as a quantitative measure of probabilistic uncertainty in the estimated model parameter given a data stream and provides a basis for determining sensor layouts that minimise this uncertainty \cite{papadimitriou2000entropy}.

Following the formulation presented in  Papadimitriou \& Lombaert \cite{papadimitriou2012effect}, consider a structural model $M(\theta) \in \mathcal{M}$ parameterised by a vector $\theta \in \mathbb{R}^{N_\theta}$ where $\boldsymbol{\theta}$ characterises either the dynamic structural behaviour or response and $\mathcal{M}$ is the set of models that constitute the digital twin and, consequently, the reinforcement learning environment.  The structural response measured at $N_{st}$ DOFs on the structure is given as $\mathbf{y} \in \mathbb{R}^{N_{st}}$ and model ($M(\theta)$) response predicted at $N_{mod}$ DOFs is given as $\mathbf{x}(\boldsymbol{\theta}) \in   \mathbb{R}^{N_{mod}}$. The relation between $\mathbf{y}$ and $\mathbf{x}(\boldsymbol{\theta})$ satisfies
\begin{equation}
    \mathbf{y} = \mathbf{L}\mathbf{x}(\boldsymbol{\theta}) + \mathbf{e}(\boldsymbol{\theta}),
\end{equation}
where $\mathbf{x}(\boldsymbol{\theta})$ represents the displacement, velocity, or acceleration response of the model, and $\mathbf{e}(\boldsymbol{\theta})$ accounts for prediction errors, including both model and measurement errors. The matrix $\mathbf{L}$ is a binary observation matrix that specifies the DOFs monitored by the sensors. In this context, $\mathbf{L}$ is considered a one-hot encoding of the state vector $s$, which defines the monitored DOFs associated with the response $\mathbf{x}(\boldsymbol{\theta})$. Based on Bayesian statistics, under a Gaussian likelihood assumption the probability density function of $\boldsymbol{\theta}$, given a set of measured data $\mathbf{D} \supset \{y_k \in \mathbf{y} \mid k = 1, \dots, n\}$, is expressed as:
\begin{equation}
p(\boldsymbol{\theta} \mid \boldsymbol{\Sigma}, \mathbf{D}) = c \frac{1}{\left( \sqrt{2\pi} \right)^n \sqrt{\det (\boldsymbol{\Sigma})}} \exp \left[ -\frac{nN_{st}}{2} J(\boldsymbol{\theta}; \boldsymbol{\Sigma}, \mathbf{D}) \right] p(\boldsymbol{\theta}),
\end{equation}
where $p(\boldsymbol{\theta})$ is the prior distribution of $\boldsymbol{\theta}$, $c$ is the marginal proportionality constant, and $J(\boldsymbol{\theta}; \mathbf{D}, \boldsymbol{\Sigma})$ quantifies the distance between measured and model-predicted response and is given as
\begin{equation}
J(\boldsymbol{\theta}; \boldsymbol{\Sigma}, \mathbf{D}) = \frac{1}{nN_{st}} \sum_{k=1}^n \left[ \mathbf{y}_k - \mathbf{Lx}_k(\boldsymbol{\theta}) \right]^\top \boldsymbol{\Sigma}^{-1} \left[ \mathbf{y}_k - \mathbf{Lx}_k(\boldsymbol{\theta}) \right].
\label{eq:response distance measure}
\end{equation}
and $\boldsymbol{\Sigma} \in \mathbb{R}^{N_{mod} \times N_{mod}}$ is the covariance matrix of the prediction error $\mathbf{e}(\boldsymbol{\theta})$ modeled as a Gaussian random vector with zero mean.

The posterior distribution provides an estimate of the uncertainty in the model parameters, $\boldsymbol{\theta}$ given the information content in the acquired data; therefore, its information entropy provides a unique scalar measure of this uncertainty and is defined by
\begin{equation}
h(\mathbf{L}; \boldsymbol{\Sigma}, \mathbf{D}) = \mathbb{E}_{\boldsymbol{\theta}}[-\ln p(\boldsymbol{\theta} \mid \boldsymbol{\Sigma}, \mathbf{D})] 
= - \int \ln p(\boldsymbol{\theta} \mid \boldsymbol{\Sigma}, \mathbf{D}) p(\boldsymbol{\theta} \mid \boldsymbol{\Sigma}, \mathbf{D}) \, d\boldsymbol{\theta}.
\label{IE}
\end{equation}
Given a sufficiently large amount of data, $(n \to \infty )$ \cite{papadimitriou2004optimal} Equation \ref{IE} can be replaced by an asymptotic approximation. The approximate information entropy is defined as
\begin{equation}
h(\mathbf{L};\boldsymbol{\Sigma},\mathbf{D}) \simeq H(\mathbf{L};\boldsymbol{\theta},\boldsymbol{\Sigma}) = \frac{1}{2}N_\theta \ln(2\pi) - \frac{1}{2}\ln[\det \mathbf{Q}(\mathbf{L};\boldsymbol{\theta},\boldsymbol{\Sigma})]
    \label{eq:entropy}
\end{equation}
where $\mathbf{Q}(\mathbf{L};\boldsymbol{\theta},\boldsymbol{\Sigma}) $  is a $N_{st} \times N_{st}$ semi-positive definite matrix known as the FIM which contains information about the uncertainty in the estimate of parameters $\boldsymbol{\theta}$ given the measured data from all monitored positions specified in $\mathbf{L}$ and is asymptotically approximated by
\begin{equation}
\mathbf{Q}(\mathbf{L};\boldsymbol{\theta},\boldsymbol{\Sigma})=\sum_{t=1}^{n}(\mathbf{L}\nabla_{\theta}\mathbf{x}_{t}(\boldsymbol{\theta}))^{T}(\mathbf{L}\boldsymbol{\Sigma} \mathbf{L}^{T})^{-1}(\mathbf{L}\nabla_{\theta}\mathbf{x}_{t}(\boldsymbol{\theta}))
    \label{eq:FIM}
\end{equation}
in which $\nabla_{\theta}\mathbf{x}_{t}(\boldsymbol{\theta})$ is the gradient of the model response with respect to its parameters $\boldsymbol{\theta}$.

In modal analysis, the response vector $\mathbf{x}(\boldsymbol{\theta})$ can be expressed as a linear combination of the mode shape, $\mathbf{x}(\boldsymbol{\theta}) = \boldsymbol{\Phi} \boldsymbol{\theta}$ where $\boldsymbol{\theta} = \boldsymbol{\zeta}  \in \mathbb{R}^m (m \leqq N_{mod})$ is the modal coordinate vector and $\boldsymbol{\Phi} \in \mathbb{R}^{N_{mod} \times m}$ is the mode shape matrix. Therefore, steering sensors to positions where the covariance of the estimated $\boldsymbol{\zeta}$ is minimal maximises the information content of the measured responses. The gradient in Equation \ref{eq:FIM} is then given as $\nabla_{\theta}\mathbf{x}_{t}(\boldsymbol{\theta}) = \boldsymbol{\Phi}$ and substituting back yields
\begin{equation}
\mathbf{Q}(\mathbf{L};\boldsymbol{\theta},\boldsymbol{\Sigma}) \equiv  \mathbf{Q}(\mathbf{L}, \boldsymbol{\Sigma}) =(\mathbf{L} \boldsymbol{\Phi})^T (\mathbf{L} \boldsymbol{\Sigma} \mathbf{L}^T)^{-1} (\mathbf{L}\boldsymbol{\Phi})
    \label{eq:FIM-phi}
\end{equation}
 as presented in Papadimitriou \& Lombaert \cite{papadimitriou2012effect}. Equation \ref{eq:FIM-phi} shows, therefore, that the FIM is only dependent on the observation matrix $\mathbf{L}$, the mode shape $\boldsymbol{\Phi}$ and the covariance matrix $\boldsymbol{\Sigma}$. It is worth noting that $\mathbf{L}$ is a non-square matrix, which prevents the trivial simplification of Equation \ref{eq:FIM-phi}.

\subsubsection{Spatial correlation of prediction error}
The spatial correlation of the prediction error is essential for the formulation of the sensor configuration reward at each learning step, as it enforces a minimum distance between sensors within the 3D space. As previously established, the prediction error $\mathbf{e}(\boldsymbol{\theta})$ is evaluated as the sum of the model and measurement error, and if independence is assumed, $\boldsymbol{\Sigma}$ takes the form
\begin{equation}
    \boldsymbol{\Sigma} = \boldsymbol{\Sigma}_{mea} + \boldsymbol{\Sigma}_{mod},
\end{equation}
where \( \boldsymbol{\Sigma}_{mea} \) represents the measurement error covariance and \( \boldsymbol{\Sigma}_{mod} \) accounts for the model error covariance. It is reasonable to assume measurement \cite{papadimitriou2012effect} errors are location-independent, resulting in a diagonal covariance matrix $\boldsymbol{\Sigma}_{mea} = c^2 \mathbf{I}$ where $c$ is a variance constant and $\mathbf{I}$ is an identity matrix. In contrast, model errors often exhibit spatial correlation between neighbouring locations due to underlying model dynamics, requiring a non-diagonal covariance matrix to capture these dependencies. The mode error covariance can then be expressed as an exponential function given as
\begin{equation}
    \Sigma_{mod,ij} = \sqrt{\Sigma_{ii} \Sigma_{jj}} \exp\left(-\frac{\delta_{ij}}{\delta}\right),
    \label{eq:distance-dependent factor}
\end{equation}
where the spatial distance between DOFs \( i \) and \( j \), denoted as \( \delta_{ij} \), is used to characterise the spatial correlation of signals, the parameter \( \delta \), known as the correlation length, quantifies the extent of spatial correlation and the auto-correlation terms \( \Sigma_{ii} \) and \( \Sigma_{jj} \) are assumed to be equal to one, representing perfect correlation at similar locations. The cross-correlation term \( \Sigma_{ij} \) spans from zero to one, where \( \Sigma_{ij} = 1 \), indicates perfect spatial correlation between sensors, and when \( \Sigma_{ij} \to 0 \), the correlation decreases as the distance \( \delta_{ij} \) between sensors increases. This relationship reflects the diminishing spatial correlation with increasing sensor separation.

However, Equation \ref{eq:distance-dependent factor} has been reported to result in suboptimal reward due to boundary case placement preference \cite{kim2024effective}. To address this we rely on a correlation function  \cite{vincenzi2017influence}, such that for the same DOFs $i,j$, the model error covariance is expressed as:
\begin{equation}
    \Sigma_{ij} = \exp\left(-\frac{\delta_{ij}}{\delta}\right) \frac{\boldsymbol{\psi}_i^\top \boldsymbol{\psi}_j}{N_M},
\end{equation}
where $\mathbf{\boldsymbol{\psi}}_i = \psi_{\textit{k},i} \mid \textit{k} = 1, \ldots, K $ and $\mathbf{\boldsymbol{\psi}}_j = \psi_{\textit{k},j} \mid \textit{k} = 1, \ldots, K $ are evaluated as:
\begin{equation}
    \psi_{k,i} = \frac{|\phi_{k,i}|}{\max(|\phi_{k,i}|, |\phi_{k,j}|)}, \quad \psi_{k,j} = \frac{|\phi_{k,j}|}{\max(|\phi_{k,i}|, |\phi_{k,j}|)},
\end{equation}
and \( \delta \) is defined as the ratio of the greatest distance across all DOFs to the total number of sensors. The terms \( \phi_{k,i} \) and \( \phi_{k,j} \) are the mode shapes at positions \( i \) and \( j \) for mode \( k \), respectively. 

\begin{figure}[htbp]
\centering
\includegraphics[width=\linewidth]{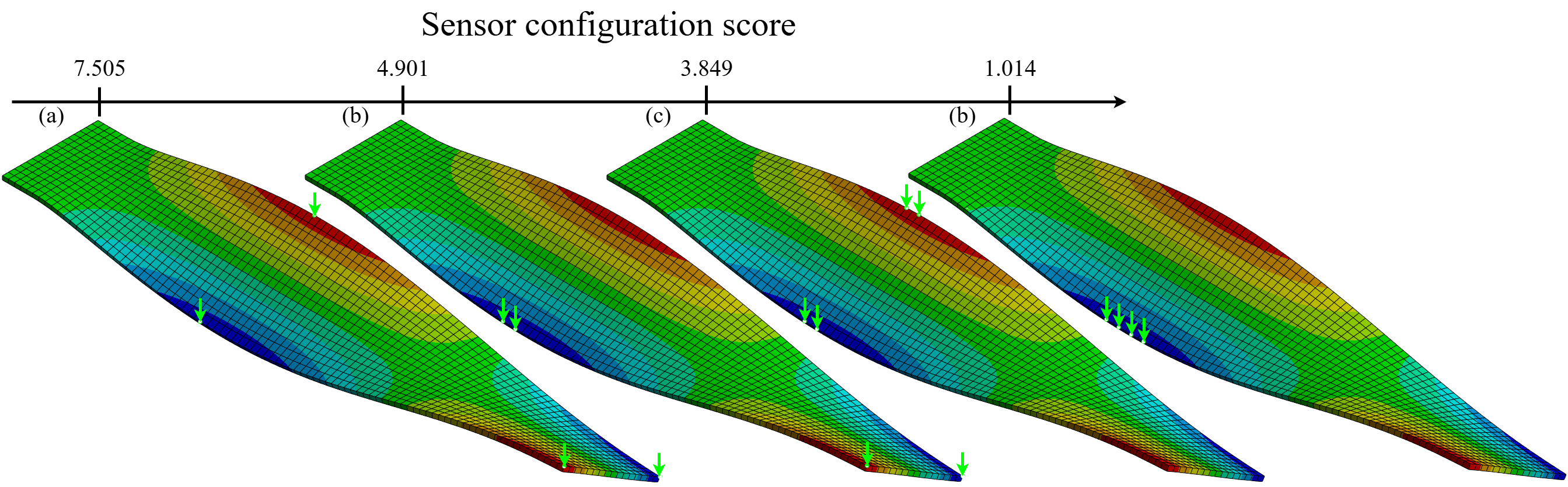}
\caption{Sensor configurations score for the  2nd torsional mode shape of a clamped cantilever structure showing (a) four sensors with an efficient spatial distribution given a high score, (b) reduced sensor configuration score due to two close sensors, (c) further reduction in sensor configuration score resulting from two pairs of close sensors, (d) poor spatial distribution of sensor with low configuration score.  The reward function takes into consideration
 the distance between sensors and penalises configuration with low spatial distribution.}
\label{Fig:reward score robustness}
\end{figure}

The terms \( \psi_{k,i}, \psi_{k,j}: \mathbb{R} \to [0,1]\) are set to unity when the mode shape component for $i$ and $j$ is zero, indicating full correlation. The product \( \boldsymbol{\psi}_i^\top \boldsymbol{\psi}_j: \mathbb{R} \to [0,K]\) provides a mode-shape-dependent weighting on Equation \ref{eq:distance-dependent factor} and \( \Sigma_{ii} \) is assumed to be unity (self-correlation). The robustness of this spatial correlation function is shown graphically in Figure \ref{Fig:reward score robustness}.

Equation \ref{eq:entropy} shows that minimising the information entropy is equivalent to minimising the $det(\mathbf{Q}(\mathbf{L},\boldsymbol{\Sigma}))$ and, we exploit the semi-positive-definite structure of $\mathbf{Q}(\mathbf{L},\boldsymbol{\Sigma})$ and instead compute the product of the squares of singular values of the Cholesky factorised $\mathbf{Q}(\mathbf{L},\boldsymbol{\Sigma})$ equation\cite{vincenzi2017influence}. The immediate reward $R$ is evaluated as the difference between consecutive sensor configurations, which we refer to as the score in subsequent sections of this paper
\begin{align*}
R &:= det(\mathbf{Q}(\mathbf{L'},\boldsymbol{\Sigma}))_{current} - det(\mathbf{Q}(\mathbf{L},\boldsymbol{\Sigma}))_{previous} \\
&= [\prod_{i=1}^{N_{mod}} \sigma_{i}^{2}]_{current} - [\prod_{i=1}^{N_{mod}} \sigma_{i}^{2}]_{previous}
\end{align*}
where $\sigma_{i}$ is the singular value. The agent receives a positive reward if a change in the position of a sensor within the current configuration results in an increase in the determinant of FIM compared to the previous configuration. 

\subsubsection{Model dynamics and discounting factor }
The model dynamics, also known as the transition probability, defines the likelihood of transitioning from one state to another given a specific action. In the context of the sensor steering problem, the model dynamics are deterministic. This means that given an action $a$ and a state $s$, the probability of transitioning to a new state $s'$ is 1. For example, if a sensor is moved in a specific direction, the new configuration of the sensors is determined with certainty. At the boundaries of the placement region, the probability of taking an action that moves a sensor outside the region is $0$, and the state $s$ remains unchanged.

The discounting factor $\gamma$ determines the importance of future rewards relative to immediate rewards. A discount factor close to $1$ implies that future rewards are nearly as important as immediate rewards, encouraging the agent to consider long-term benefits. Conversely, a discount factor close to $0$ places more emphasis on immediate rewards, potentially leading to shortsighted decisions. In the context of the sensor steering problem, the discounting factor helps balance the trade-off between immediate improvements in a single sensor placement and the long-term goal of optimising the overall sensor configuration. By appropriately tuning the discount factor, whether through optimization or heuristics, the agent can be directed to make decisions that achieve the optimal final configuration.

\subsection{Reinforcement learning}
Reinforcement learning is based on an MDP formulation and involves having an agent interact with an environment to learn a policy which maximises a specified reward \cite{barto2021reinforcement}. The sensor steering problem can be viewed as a sequential decision process that involves decisions made by an agent $\pi_t(a,s): Pr\{ A_t = a | S_t = s\} $ about where to move the sensors given the current state of the digital twin. For a robust digital twin framework, it is valuable to make these decisions with a consideration of uncertainty or some measure of risk rather than just expectations. A natural algorithm choice that presents these features is a value-based distributional reinforcement learning \cite{bellemare2017distributional}. Here the aleatoric (intrinsic) uncertainty due to the stochasticity of the policy and environment is modelled by the return distribution. More precisely, we utilise the deep Q-network algorithm (DQN) called Rainbow \cite{hessel2018rainbow} which combines significant algorithmic improvements of the original DQN. Rainbow is formulated based on an equivalent distributional Bellman's equation given as
\begin{equation}
    Z(s, a) \overset{D}{=} R(s, a) + \gamma Z(S', A')
\end{equation}
where $Z(s, a)$ is the return distribution whose expectation is the value function $Q(s,a)$ which is not related to Equation \ref{eq:FIM-phi} and is given as
\begin{equation}
   Q(s,a) := \mathbb{E} Z(s, a) = \mathbb{E} \biggr[ \sum_{t=0}^{T} \gamma^{t}R(s, a)\biggr]
\end{equation}
and $R$ is a random variable representing the reward. The learning involves minimising the Kullback-Leibler divergence between the online distribution $Z_t$ and the target distribution $Z_t^{(q)}$, given as
\begin{equation}
   D_{KL}(\Psi_{z}Z_{t}^{(q)}||Z_{t})
\end{equation}
where $\Psi_{z}$ is the projection onto $Z$ since the online and target distributions have disjointed support. The return distribution is parameterised by a categorical distribution over a fixed number of points called atoms, which are equidistant and bound by $[V_{min}, V_{max}]$ representing the canonical returns.

\section{Proposed framework}\label{sec:Proposed Framework}
\begin{figure}[htbp]
  \centering
    \includegraphics[width=1\textwidth]{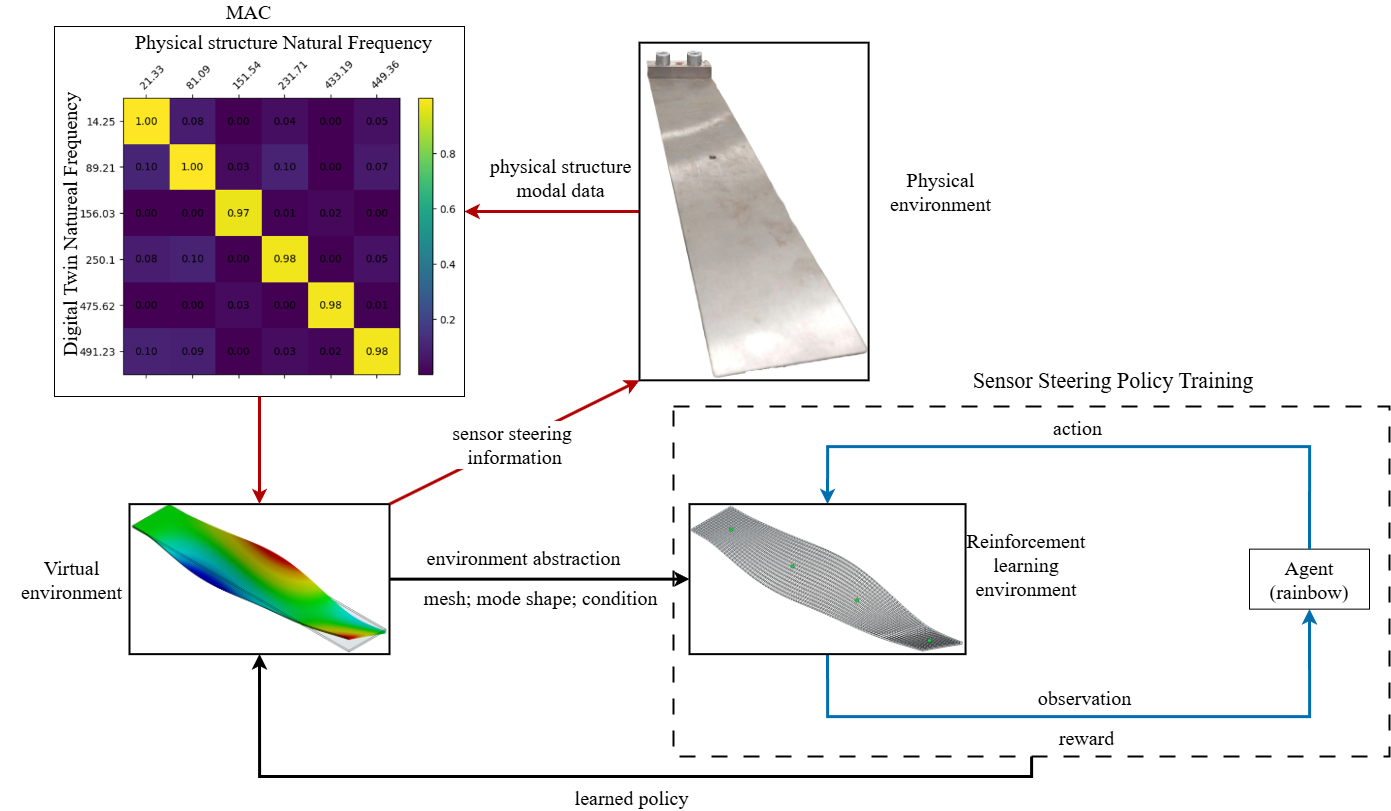} 
  \caption{Graphical representation of interaction between physical and digital space. The red cycle indicates the virtual-to-physical and physical-to-virtual interaction; the black cycle shows the interaction between the virtual and reinforcement learning environments and the blue cycle is the agent-environment interaction.}
  \label{Fig: DT-Sensor Steering integration}
\end{figure}

This section describes how the proposed strategy fits into the digital twin paradigm. The learning environment is considered an abstraction of the virtual environment defined by the current structural condition, mesh information, sensor details, and mode shape as shown in Figure  \ref{Fig: DT-Sensor Steering integration}. With this data, a custom RL environment is set up using the OpenAI Gymnasium environment framework \cite{towers2024gymnasium}. An optimal policy is learnt by steering all available sensors within the environment to explore various sensor placement configurations and thereby find the shortest path to the optimal positions. At each environment initialisation (reset), all available sensors are evenly distributed along the longitudinal line of symmetry of the structure, representing a heuristic configuration upon which the agent improves (see Figure  \ref{Fig: DT-Sensor Steering integration}). The trained agent is deployed to the virtual space, where it adapts the position of the sensors given a known virtual environment state, and finally, this information is made available to the physical space.

In the physical space, a typical sensor steering step involves sensors that are capable of moving along tracks, such as rails. Given information from the digital twin, these sensors are dynamically repositioned along the rails to new positions of interest. One notable possible application of this approach is in the monitoring of large linear structures such as bridges. By installing rails along the length of the bridge, sensors could be systematically steered to various locations as needed. This strategy not only optimises data collection but also results in a significant reduction in the number of installed sensors. Instead of deploying a large number of fixed sensors across the entire structure, a smaller set of mobile sensors could cover multiple points of interest, thereby reducing costs and simplifying maintenance.

\subsection{Implementation details}\label{sec:Implementation details}
This paper focuses on the sensor steering strategy, assuming that an instance of the digital twin with a good representation of the physical asset is available. Therefore, we do not include a continuous validation and updating process. The physical structure is a steel cantilever plate with the following dimensions: 447 mm in length, 76.2 mm in width, and 3 mm in thickness. The plate is clamped at one end at a depth of 24 mm from the end. A finite element model of the physical structure with an element size of $5$ mm and similar boundary conditions is constructed, and the computed modal assurance criterion (MAC) is shown in Figure  \ref{Fig: DT-Sensor Steering integration} (top left). The MAC demonstrates a high correlation in the mode shape, which is crucial for computing the learning reward. The MAC values also validate that our simulation model accurately represents the physical structure. 
To build the virtual environment, the simulation model is scripted into a sequence of commands using the Ansys Pythonic API (PyMechanical). Data from the virtual space is extracted and used to define the custom reinforcement learning (RL) environment. Figure  \ref{Fig: Evolution of sensing strategy} shows the RL environment search region defined by the candidate sensor location. Different options were considered for handling edge location actions. Initially, we defined agent actions outside the search region as episode termination actions, which resulted in resetting the environment. However, this setup proved to be inefficient and led to short learning episodes. Subsequently, we defined actions outside the search region as null actions, meaning the next state is the same as the previous state. This approach proved to be effective for the learning process. Finally, the Rainbow agent is implemented using the Tianshou Python package \cite{tianshou}. The environment configuration scores are normalised to a maximum of $1$ and $\delta = 0.42981$ which is the longest length across the cantilever plate placement region. The DRL agent hyperparameters are similar to values in Hessel et al.\cite{hessel2018rainbow} with the exception of $\gamma = 0.9$ and target network period $= 3200$  while the learning rate, priority exponent, and multi-step are case specific. For the exploration strategy, we adopt an $\epsilon-$greedy strategy but anneal $\epsilon$ to $0.01$ in the first 250k steps. An open-source implementation of the proposed framework reproducing our experiment is available at \href{https://github.com/Collins-Ogbodo/DT-SS}{https://github.com/Collins-Ogbodo/DT-SS}.

\section{Results}\label{sec:Results}
In this section, we demonstrate the effectiveness of our proposed framework, described in the \hyperref[sec:Problem Formulation]{Problem Formulation} section, when applied to a cantilever plate structure. We present two case studies: the first focuses on sensor steering in a structural damage severity setting (severity of the number and locations of damage), and the second on damage localisation. Considering the complexity of the placement region as indicated by $1462$ candidate sensor locations and their combinations given the number of deployed sensors, we focus on the first three vibration modes, a maximum of four deployed sensors, and a library of three damage conditions similar to the model library presented in Kapteyn \& Willcox\cite{kapteyn2022data} in both case studies, to guarantee the tractability of the problem. Damage is simulated by the addition of point masses, which result in minimal changes in the first mode shape and corresponding significant changes in higher mode shapes. These case studies illustrate the practical application of our framework and its potential to enhance the accuracy and reliability of digital twins in dynamic environments.

\subsection{Adapting based on damage severities}\label{subsec:Damage Severity}
\begin{figure}[htbp]
  \centering
    \includegraphics[width=1\textwidth]{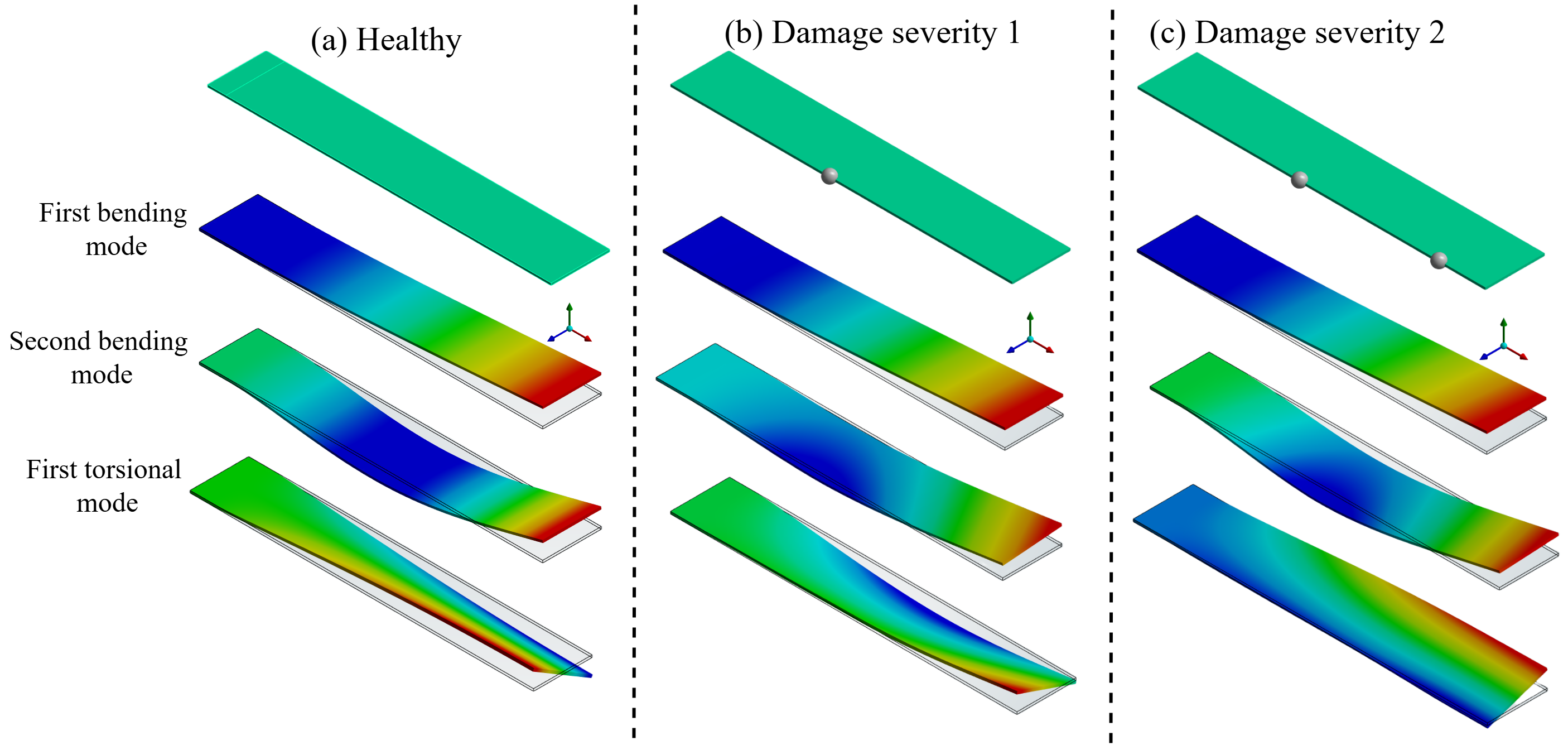} 
  \caption{The first three mode shapes of different structural conditions of a cantilever plate showing (a) the healthy condition, (b) damage condition 1 simulated by a point mass of $0.7Kg$ and (c) damage condition 2 simulated by two point masses of $0.7Kg$ on the same edge.}
  \label{Fig: Damage severity}
\end{figure}

Our first consideration is a damage severity problem. The goal, therefore, is to learn a policy capable of adapting sensor positions with increasing degrees of damage. We consider two different damage severities, in addition to the healthy condition. The healthy condition represents the baseline state of the cantilever plate without any structural issues. The two damage conditions simulate different scenarios where the plate experiences structural damage. These conditions are illustrated in Figure  \ref{Fig: Damage severity}.

Setup:
\begin{itemize}
    \item \textbf{Healthy condition}: The cantilever plate is in its original, undamaged state (Figure  \ref{Fig: Damage severity}a).
    \item \textbf{Damage severity 1}: A point mass of 0.7 $Kg$ simulates a moderate level of damage at the edge of the plate (Figure  \ref{Fig: Damage severity}b).
    \item \textbf{Damage severity 2}: Two point masses of 0.7 $Kg$ each, which simulate an increased level of damage on the same edge side of the plate (Figure  \ref{Fig: Damage severity}c).
\end{itemize}

\begin{figure}[htbp]
  \centering
    \includegraphics[width=1.0\textwidth]{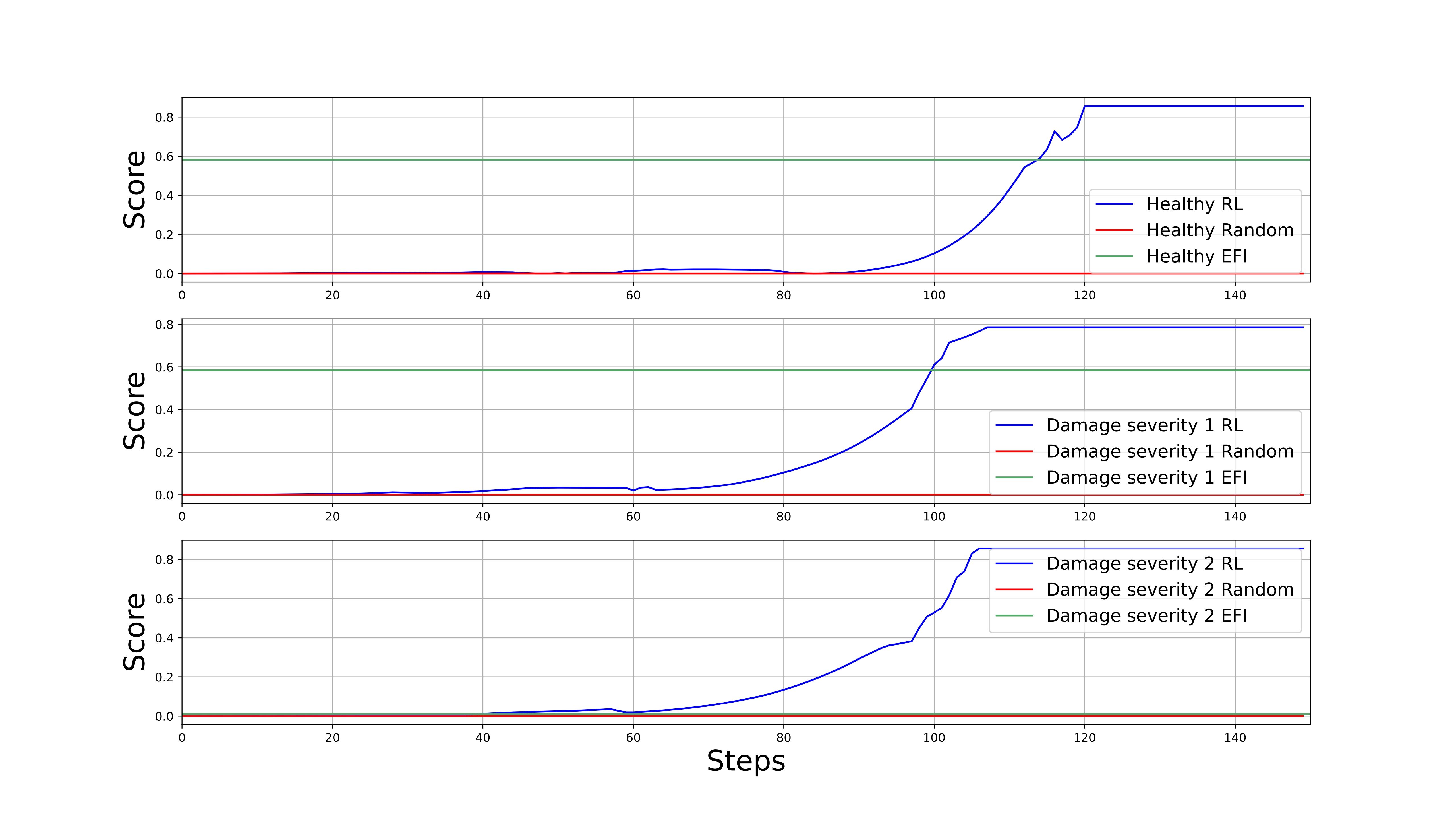} 
  \caption{Agent condition-specific evaluation showing score for healthy (top), damage severity 1 (middle) and damage severity 2 (bottom)}
  \label{Fig: Damage severity evaluation}
\end{figure}

For this case study, the environment is initialised with three sensors as described in the \hyperref[sec:Proposed Framework]{Proposed Framework} section. The agent is trained using the DRL agent with case-specific hyperparameters of learning rate $=6.25 \times 10^{-5}$, priority exponent $=0.5$, multi-step $=3$, and a fully connected multi-layer perceptron (MLP) with two layers consisting of $256$ units and $128$ units, respectively. To demonstrate the condition-specific performance, Figure  \ref{Fig: Damage severity evaluation} presents a comparative analysis of the trained agent against an effective independence (EFI) methodology \cite{kim2024effective} with the same error covariance matrix and a random agent as baselines. The results indicate that the trained agent significantly outperforms both baselines within each condition. The trained agent achieves a score above $0.85$ in both the healthy and damage severity $2$ conditions within $120$ and $105$ steps, respectively. In the case of damage severity $1$, it reaches a score of approximately $0.79$, still exceeding the performance of the baselines.

\begin{figure}[htbp]
  \centering
    \includegraphics[width=1.0\textwidth]{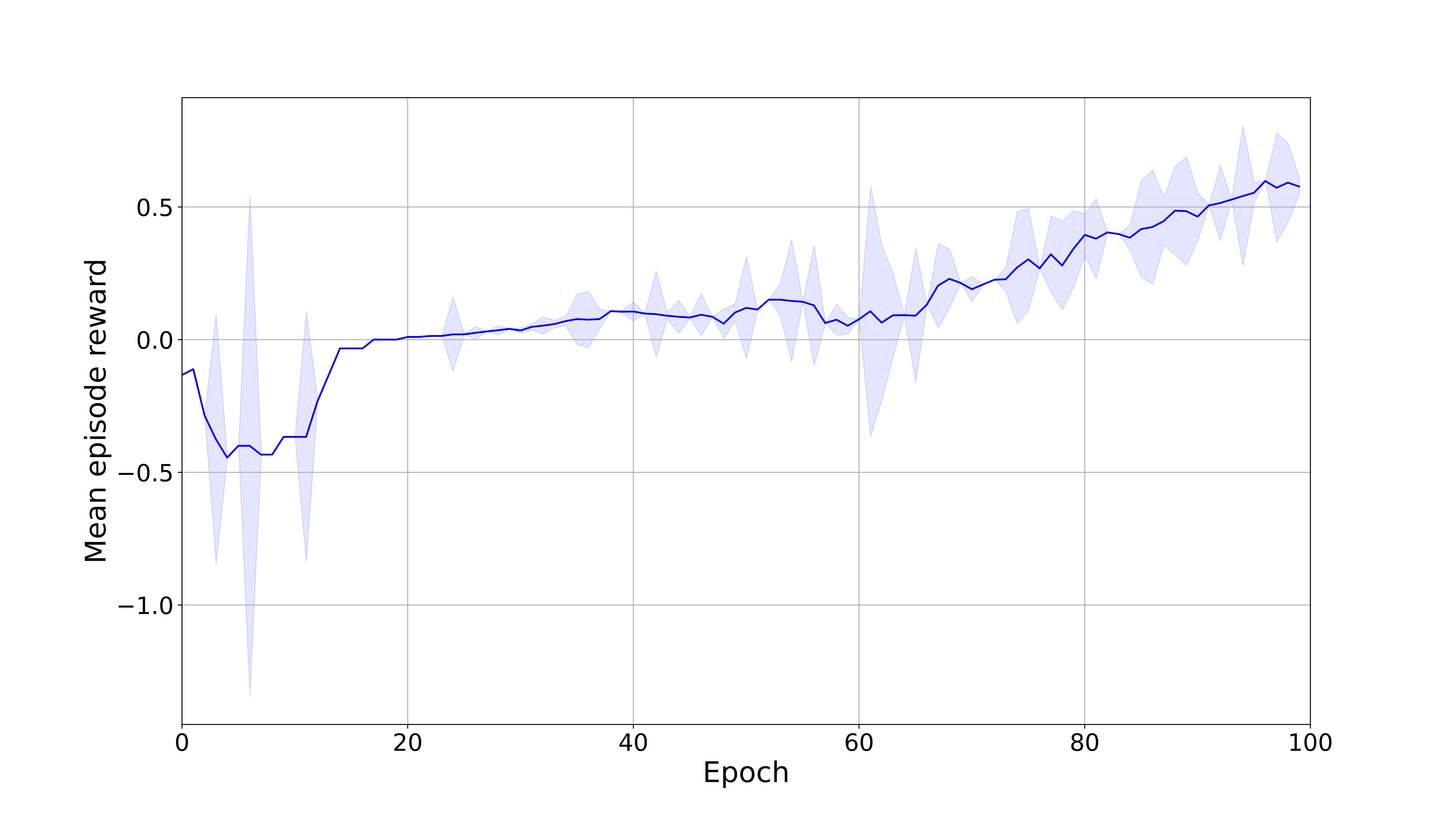} 
  \caption{Mean episode reward averaged over three test episodes for healthy and damage severity 1 and 2 with standard deviation (shaded region). Curve is smoothed with a moving average of 10 points}
  \label{Fig: Damage severity mean episode reward}
\end{figure}

\begin{figure}[htbp]
  \centering
    \includegraphics[width=1.0\textwidth]{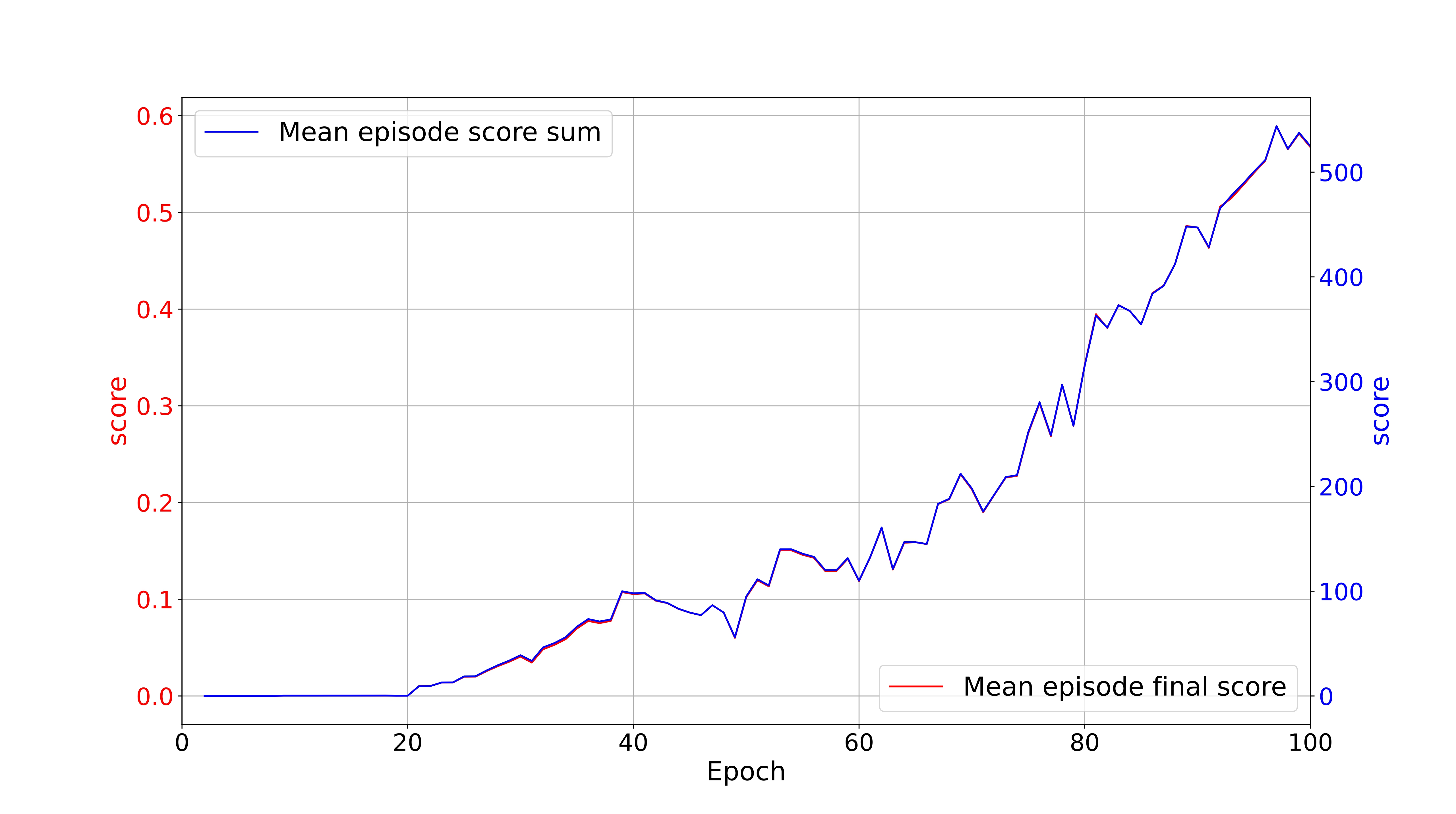} 
  \caption{ Score metric averaged over three test episodes per epoch for healthy and damage severity 1 and 2: episode score sum (red) and episode final score (blue). Similarity of plots shows the agent's performance in following the optimal path to the optimal location. Curve is smoothed with a moving average of 10 points}
  \label{Fig: Damage severity score}
\end{figure}

To achieve this performance, we further describe the agent training approach. Figure \ref{Fig: Damage severity mean episode reward} shows the agent's mean episode reward performance averaged over three test episodes per epoch. This indicates the agent's capacity to learn an optimal sensor steering policy across all three structural conditions. The shaded region represents the agent-environment interaction standard deviation. Each training epoch consists of 10,000 steps, and each environment episode terminates after 1,000 steps. After each environment termination, a random structural condition is sampled for the next episode to enable the policy to generalise effectively to all structural conditions. This approach is similar to domain randomisation technique for sim-to-real problems within the robotic community\cite{tobin2017domain}. The final score performance, which is the final score at the end of each episode averaged over test episodes, is shown in Figure \ref{Fig: Damage severity score}. The final score started with a low value at the early stages of training and increased to a maximum of $0.8039$, demonstrating its effectiveness in accurately repositioning sensors.

Finally, we also show that the agent learns the optimum path to the position that maximises the configuration score, as described in the \hyperref[subsec:Reward Function]{Reward Function} section. Within a test episode, the agent first steers the sensors to their optimal positions and then maintains this final configuration for the remaining steps. Consequently, a typical episode score trajectory is dominated by the final score, making the plot of the sum of a test trajectory score approximately similar to the final score but scaled by the length of the test episode. This proximity is desirable and indicates that in each episode, only a few initial steps are taken by the agent to reach the optimal positions. This similarity property is shown in Figure  \ref{Fig: Damage severity score}.


\subsection{Adapting to different damage locations}\label{subsec:Damage Location}
\begin{figure}[htbp]
  \centering
    \includegraphics[width=1\textwidth]{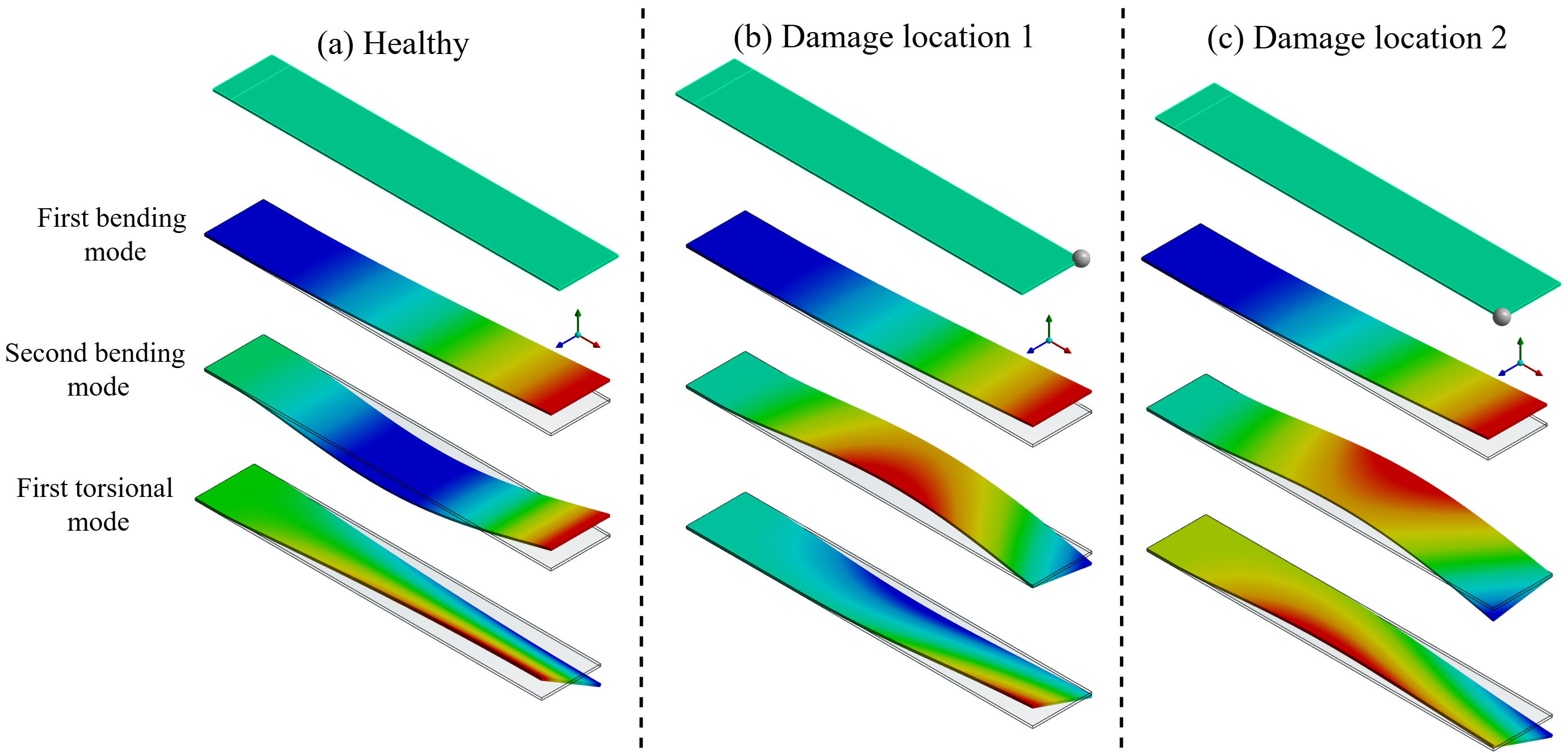} 
  \caption{The first three mode shapes of different structural conditions of a cantilever plate showing (a) the healthy condition, (b) damage condition 1 simulated by a point mass of $0.2Kg$ at the right vertex, and (c) damage condition 2 simulated by a point mass of $0.2Kg$ at the left vertex.}
  
  \label{Fig: Damage Location}
\end{figure}

In this case study, we demonstrate the efficacy of the proposed framework in adapting sensors to damage at different locations of a structure. Similar to the previous case, we consider two different damage conditions, in addition to the healthy condition. The goal, therefore, is to learn a policy capable of adapting sensor positions given that damage at specific locations have altered the structure's dynamic behaviour (see Figure \ref{Fig: Damage Location}).

To provide a comprehensive analysis, we established the following conditions for the cantilever plate structure:
\begin{itemize}
    \item \textbf{Healthy condition} (Figure  \ref{Fig: Damage Location}a).
    \item \textbf{Damage conditions 1 and 2}: A point mass of 0.2 $Kg$ simulates damage at the two vertices locations of the plate, respectively (Figure  \ref{Fig: Damage Location}b,c).
\end{itemize}

\begin{figure}[htbp]
  \centering
    \includegraphics[width=1.0\textwidth]{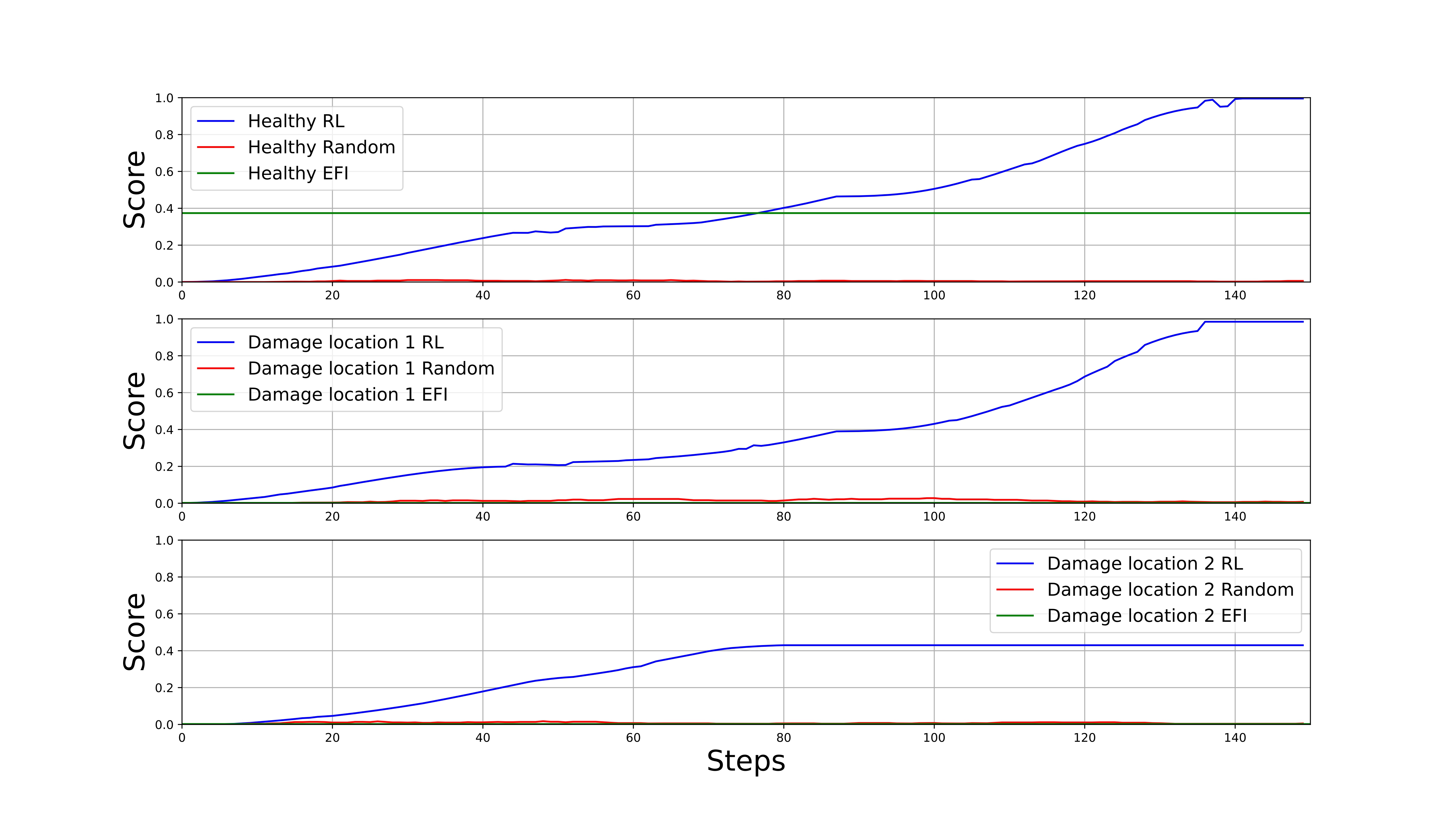} 
  \caption{Agent condition-specific evaluation showing score for healthy (top), damage location 1 (middle) and damage location 2 (bottom)}
  \label{Fig: Damage Location evaluation}
\end{figure}

\begin{figure}[htbp]
  \centering
    \includegraphics[width=1.0\textwidth]{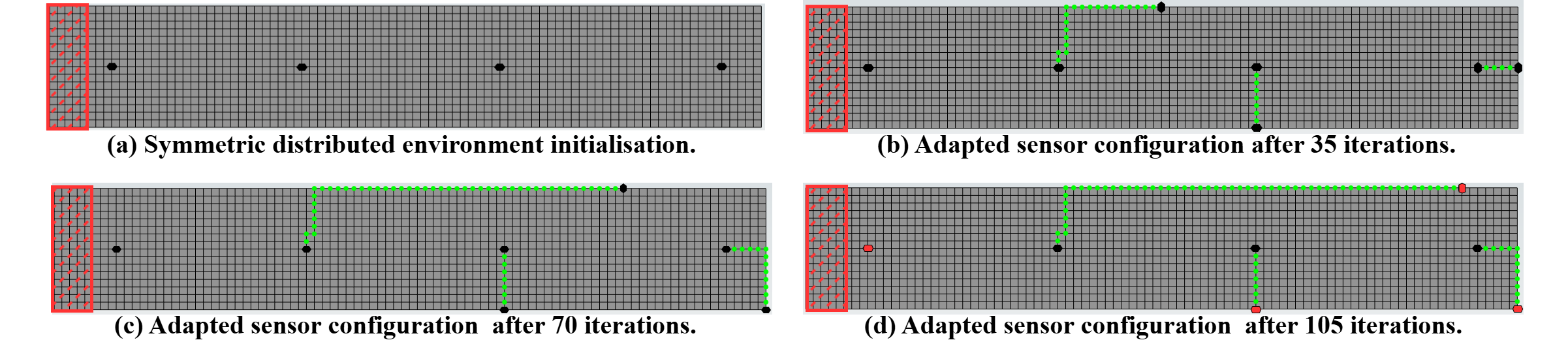} 
  \caption{Trained agent adapts sensor configuration for damage location 2 condition showing agents action path from (a) initialisation and after (b) $35$ steps, (c) 70 steps, and (d) 105 steps. The agent performance in this condition is attributed to the first sensor being fixed throughout the agent training process, as depicted by the final sensor positions (red marker).}
  \label{Fig: Damage Location path}
\end{figure}

In this case, we use four sensors, and the agent is hyperparameterised with a learning rate of $1.25 \times 10^{-4}$, a priority exponent of $0.7$, multi-step parameter of $5$ and MLP with two layers of $128$ units each. The agent is then trained using a similar approach as the previous case but for 200 epochs to ensure comprehensive learning and generalisation. We now evaluate the condition-specific performance of the agent. Figure  \ref{Fig: Damage Location evaluation} shows the performance compared to baselines described in the \hyperref[subsec:Damage Location]{Damage Location} section. The trained agent outperforms the baselines and finds the optimal location after less than $150$ steps in the healthy condition and damage location 1 condition. In the final condition (damage location 2), the agent achieves suboptimal results of $0.4$, although still better than the baselines. The agent's action trajectory in this condition is visualised in Figure \ref{Fig: Damage Location path}, and it can be seen that the first sensor remains unchanged, indicating that the agent during training learned nothing about it. This can be attributed to the increase in learning complexity introduced by one extra sensor.

\begin{figure}[htbp]
  \centering
    \includegraphics[width=1.0\textwidth]{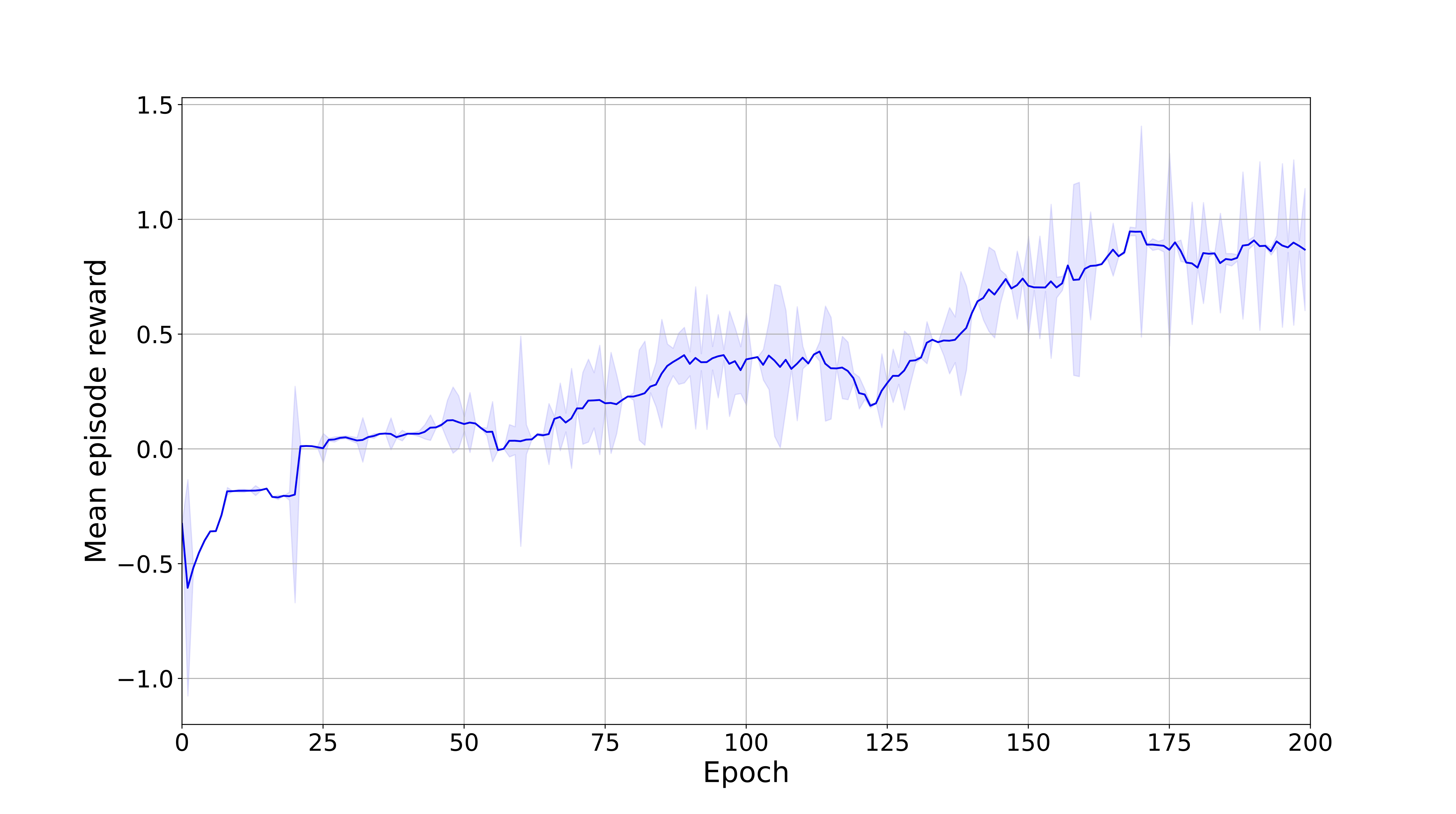} 
  \caption{Mean episode reward averaged over three test episodes for healthy and damage locations 1 and 2 with standard deviation (shaded region). Curve is smoothed with a moving average of 10 points}
  \label{Fig: Damage Location mean episode reward}
\end{figure}

\begin{figure}[htbp]
  \centering
    \includegraphics[width=1.0\textwidth]{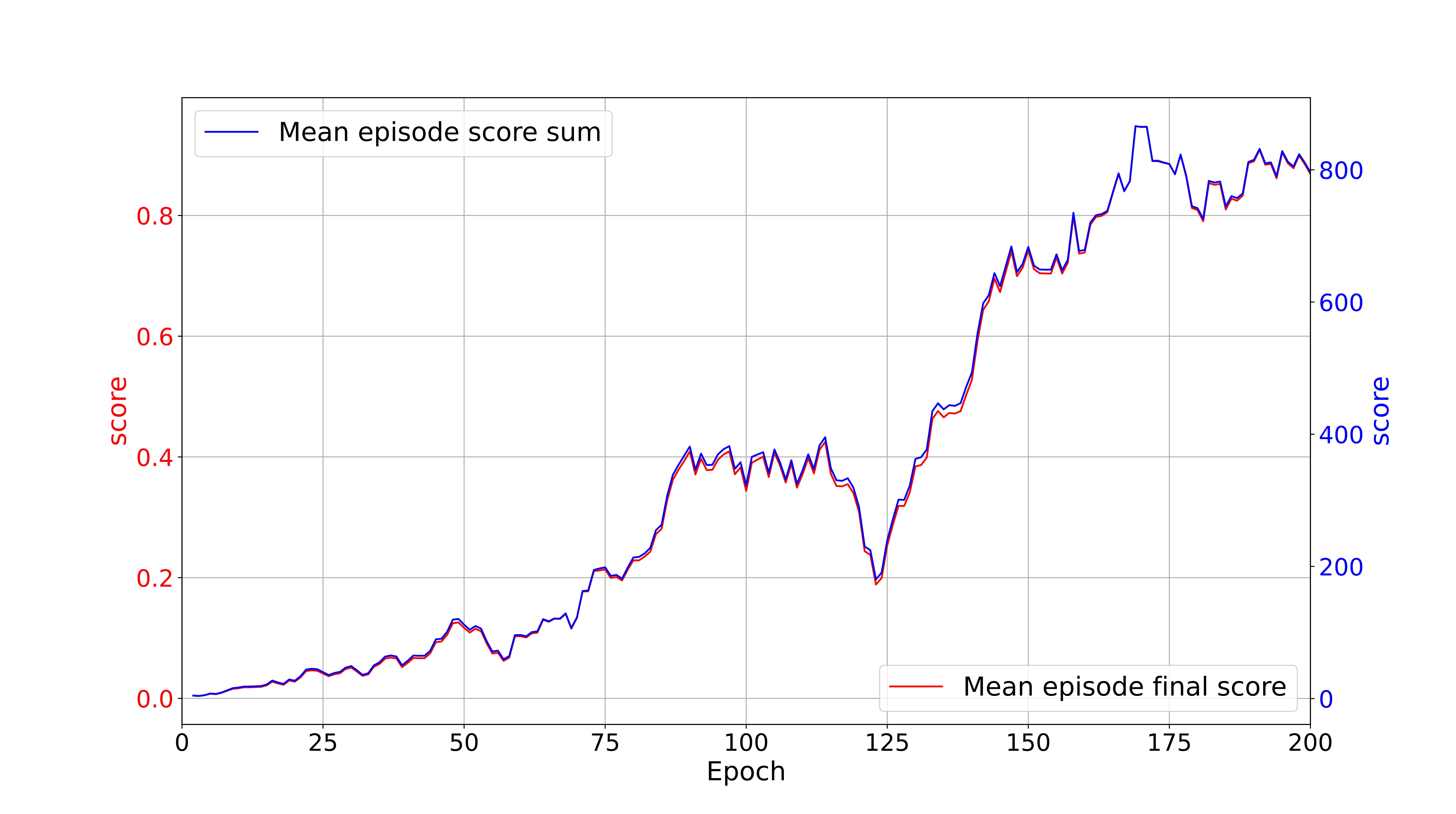} 
  \caption{Score metric averaged over three test episodes per epoch for healthy and damage locations 1 and 2: episode score sum (red) and episode final score (blue). Similarity of plot shows the agent's performance in following the optimal path to the optimal location. Curve is smoothed with a moving average of 10 points}
  \label{Fig: Damage Location score}
\end{figure}

During the training process, the agent's performance was also evaluated based on several metrics. The primary performance metric (mean episode reward), illustrated in Figure \ref{Fig: Damage Location mean episode reward}, shows that the agent achieved a maximum mean reward of $0.9982$ which was reached at the final training epoch, indicating consistent learning outcomes. In addition to the reward perfomance, we also assessed the agent's performance score and the trajectory towards optimal sensor positioning. As shown in Figure \ref{Fig: Damage Location score}, the agent achieved a maximum score of $0.9839$, demonstrating its effectiveness in accurately repositioning sensors. Figure \ref{Fig: Damage Location score} also shows a degradation in the performance of the agent between epochs 100 and 125; from our observation, this is due to increased sampling of the damage severity $2$ condition during environmental initialisation, which results in suboptimal performance. The path to optimal sensor position performance is detailed in Figure \ref{Fig: Damage Location score}, which, as explained previously, is similar to the final score plot and whose similarity indicates optimal path identification performance.

These cases demonstrate the practical application of the proposed framework in adapting the sensor locations to structural damage. The results serve as a benchmark for evaluating strategies aimed at dynamically adjusting sensor positions based on real-time feedback for timely identification and assessment of potential issues.

\section{Conclusion}\label{sec:Conclusion}
In this study, we have presented a novel approach for dynamic data acquisition in digital twins by employing adaptive sensor steering strategies based on deep reinforcement learning (DRL). This work is a first step towards adaptive sensing for improved structural monitoring. The proposed framework is motivated by the need for continuous assimilation of informative data relevant to the digital twin at various stages of its operational lifespan, a requirement for which existing optimal sensor placement techniques are not designed for. The problem is formulated as a Markov decision process and implemented with a value-based distributional reinforcement learning agent. We present a new custom reinforcement environment architecture with different damage scenarios extracted from the digital twin, and an agent is trained to act optimally by adapting the sensor positions when an observed damage condition is encountered. The trained agent is deployed in the digital twin to guide data acquisition. The proposed framework was validated through two cases on a cantilever plate structure under various conditions, including healthy and damaged conditions. Our results demonstrate the capability of the DRL-based policy to adaptively reposition sensors, hence improving the accuracy and reliability of the digital twin's representations. Both cases assume that an updated digital twin of the physical asset is available and therefore ignore the continuous updating process, which is essential in practice and should be integrated. The proposed framework can also be used for acquiring relevant data at the design stage of a digital twin (see Figure  \ref{fig:Digital Twin Life cycle}).

The proposed approach faces two key limitations: \textbf{(i)} the performance of the agent degrades significantly with the increasing number of modes considered, sensors available, and damage cases integrated. This is the curse of dimensionality. \textbf{(ii)} Similar to  Kapteyn \& Willcox \cite{kapteyn2022design}, we assume the damage states are known and fixed, which is impractical in many applications. However, the proposed methodology presents an opportunity to learn and interpolate between conditions seen during training.

A key observation during this study is the sensitivity of agent performance to the exploration strategy. Therefore, future work will focus on developing an exploration strategy specific to the problem, which allows for expedited learning and improved performance at higher levels of difficulty. Additionally, the decision-making process in the current study is based on the expectation of the value distribution. Exploring the decision-making process using other metrics is an area of future interest.

\section{Acknowledgements}\label{sec:Acknowledgements}
The authors gratefully acknowledge UK Research and Innovation (UKRI) and Siemens Digital Industries Software for their support under an EPSRC Industrial Case Award (ref. 2756020). David Wagg would like to acknowledge the support of the Alan Turing Institute and UKRI via EP/Y016289/1. Collins Ogbodo would also like to thank Rea Nkhumise for the insightful conversation that proved valuable to this work. For the purpose of open access, the author(s) has/have applied a Creative Commons Attribution (CC BY) licence to any Author Accepted Manuscript version arising.

\bibliographystyle{unsrt}
\bibliography{references}  






\end{document}